\documentclass[paper]{geophysics}
\usepackage{amsmath}
\usepackage{makecell}
\usepackage{graphicx}

\begin{document}
\title{Ground-roll Separation From Land Seismic Records Based on Convolutional Neural Network}
\renewcommand{\thefootnote}{\fnsymbol{footnote}}

\address{
\footnotemark[1]{
EasySignal Group, Department of Automation, Tsinghua University, Beijing, P.R.China
}
\footnotemark[2]{
Shengli Oilfield Geophysical Research Institute of Sinopec, Dongying, Shandong Province, P.R.China \\
}
}

\author{Zhuang Jia \footnotemark[1], Wenkai Lu \footnotemark[1], Meng Zhang \footnotemark[2], Yongkang Miao \footnotemark[2] }

\lefthead{Z. Jia \emph{et. al.}}
\righthead{Ground-roll separation based on CNN}

\maketitle

\begin{abstract}
Ground-roll wave is a common coherent noise in land field seismic data. This Rayleigh-type surface wave usually has low frequency, low apparent velocity, and high amplitude, therefore obscures the reflection events of seismic shot gathers. Commonly used techniques focus on the differences of ground-roll and reflection in transformed domain such as $f-k$ domain, wavelet domain, or curvelet domain. These approaches use a series of fixed atoms or bases to transform the data in time-space domain into transformed domain to separate different waveforms, thus tend to suffer from the complexity for a delicate design of the parameters of the transform domain filter.  To deal with these problems, a novel way is proposed to separate ground-roll from reflections using convolutional neural network (CNN) model based method to learn to extract the features of ground-roll and reflections automatically based on training data. In the proposed method, low-pass filtered seismic data which is contaminated by ground-roll wave is used as input of CNN, and then outputs both ground-roll component and low-frequency part of reflection component simultaneously. Discriminative loss is applied together with similarity loss in the training process to enhance the similarity to their train labels as well as the difference between the two outputs. Experiments are conducted on both synthetic and real data, showing that CNN based method can separate ground roll from reflections effectively, and has generalization ability to a certain extent. 
\end{abstract}

\section{Introduction}
In land seismic data processing area, ground-roll wave, a commonly visible Rayleigh-type wave which propagates within the low-velocity surface layers, sometimes covers a large part in the seismic data and conceals the information contained in the reflected events (\cite{saatcilar_method_1988}). Various attempts has been tried to attenuate the ground-roll wave in seismic gathers. One possible strategy is to attenuate ground-roll in the data acquisition stage, for example: by summation of CMP traces or polarization filtering process (\cite{morse_groundroll_1989,shieh_ground_1990}). However this strategy has restricted effectiveness to satisfy our requirement of small ground-roll energy remaining and slight deterioration of reflected events. Therefore, signal processing techniques for ground-roll attenuation after data acquisition are essential and necessary.

Currently, the most common techniques for ground-roll suppressing task usually take advantage of the difference between ground-roll and reflections in frequency domain or frequency-wavenumber domain of 2-D trace gathers. It is known as a prior knowledge that ground-roll wave has the properties of high amplitude, low frequency, low apparent velocity and low phase velocity. The basic intuition in most of the techniques is to transform the seismic shot gather contaminated by ground-roll wave into a certain domain where ground-roll and reflections are separated as much as possible due to their difference in frequency and velocity, so that a pre-designed or adaptive mask can be applied to extract the reflection part or ground-roll part respectively, then via inverse transform the reflections and ground-roll can be reconstructed independently. Based on this thought, a number of methods with different transform domain has been proposed. The early proposed techniques usually mainly focus on frequency domain or $f-k$ domain, such as simple high-pass filtering, $f-k$ dip filtering, wide-band velocity filtering in $f-k$ domain (\cite{embree_wideband_1963}), $f-k$ filtering using Hartley transform (\cite{gelisli_f-k_1998}). These methods may cause elimination of low frequency part of reflections while attenuating ground-roll. In addition, several other transforms are also utilized to separate reflections and ground-roll, for example, wavelet transform using 1-D wavelet or 2-D physical wavelet (\cite{deighan_groundroll_1997,wang_data_2012,zhang_physical_2003}), curvelet transform (\cite{herrmann_non-parametric_2008,liu_ground_2018,naghizadeh_ground-roll_2018}), and Radon transform (\cite{trad_hybrid_2001}) is applied to discriminate ground-roll and reflections by their linear and hyperbolic event curvatures. 

All the above methods can effectively attenuate ground-roll from land seismic data to different extent. However, these techniques use fixed \textquotedblleft atom\textquotedblright{} or \textquotedblleft basis\textquotedblright{} to decompose and reconstruct the seismic signal and delicately designed masks to separate signals, thus requires relatively much work on the analysis of data and choosing of an apropriate parameter set. Moreover, such general transform method with fixed dictionary of atoms which can be applied to many quite different tasks may not perform the optimum result on this specific task. 

Recently, convolutional neural network (CNN) has become widely used in image processing area and won the state-of-the-art performance in many related problems, such as image recognition (\cite{He2016Deep, Simonyan2014Very, Szegedy2016Rethinking}), object detection (\cite{Ren2015Faster, Redmon2017YOLO9000, Liu2016SSD}), and super-resolution (\cite{cheong_deep_2017, huang_wavelet-srnet:_2017, Ren2018CT}). Moreover, in the seismic data processing area, many CNN based methods have also been proposed to deal with tasks such as denoising, interpolation, and seismic image enhancement(\cite{liu2018random, si2018random, li2018deep, wang2018intelligent, halpert2018deep}), and received promising results.

All the above achievements are based on the ability of CNN to extract the features of images and learn the \textquotedblleft kernel\textquotedblright{} or \textquotedblleft filter\textquotedblright{} to represent the image automatically in supervised training process (\cite{masci_stacked_2011}). For the task of ground-roll separation, as we conduct it in the t-x domain, the key is to try to utilize the differences of reflections and ground-rolls. In t-x domain, these differences are mainly in the 2-D morphological features. In order to extract the 2-D features of reflections and ground-rolls, and utilize them to reconstruct the two components respectively, we use CNN model which has the ability to extract 2-D features of reflections and ground-roll. Moreover, instead of applying a popular CNN architecture directly in the seismic data processing field as the above mentioned methods,  we take the task of ground-roll noise attenuation as a separation problem considering the inspiring results in monaural source separation task using CNN based methods (\cite{huang_deep_2014, Huang2015Joint, Grais2018Single}), where we need to separate the ground-roll noise and reflections apart from the  mixture of recorded seismic data. In this paper, we proposed a method using CNN based model to separate ground-roll and reflections apart. We train our CNN model with both similarity loss (i.e. mean square error between output and the given train label to supervise training) and discriminative loss (constraint on the difference between two outputs). Then the trained CNN model is applied to ground-roll contaminated seismic shot gather to separate the mixture into ground-roll and reflections. Experiments show that our CNN model can reach a satisfying result on both synthetic data and real field data, and the generalization test on synthetic data with properties different from train dataset (such as sample rate, ground-roll frequency band and noise level), shows that the trained model has the generalization ability to a certain extent.

The remainder of the paper is organized as follows. In Section 2, we demonstrate the steps of our method for ground-roll separation and describe the CNN model architecture. Section 3 shows the experiment results with both the synthetic data example and real seismic data example. In synthetic data example we also illustrate the generalization ability of our method via testing on synthetic data with different properties (time and space sample rate, magnitude and frequency dispersion of ground-roll, and noise level). Section 4 draws the conclusion of our research and discusses the probable improvement in the future work.

\section*{Methodology}
\subsection{Basic idea of the method}

The flowchart of proposed method is shown in Fig. 1. Detailed illustration of our method is as follows. Firstly, we denote the mixture of reflections and ground-roll as $m$ , and reflections and ground-roll as $r$
and $g$ respectively. Then $m$ is the summation of $g$ and $r$:

\vspace{-10px}

\begin{equation}
m=g+r
\end{equation}

As is known to us that the ground-roll wave occupies merely the low frequency part of the mixed signal, that is, the ground-roll has a bandwidth lower than a certain maximum frequency, usually 30 Hz or even lower. Therefore we apply a low-pass frequency filter to $m$ , which has a cutoff frequency larger than the maximum frequency of $g$ , and then we can get the low-frequency component of $m$, denoted as $m_{low}$ . Then we subtract $m_{low}$ from $m$ to get $m_{high}$. Because of the limited bandwidth of $g$ , $m_{high}$ contains no energy of ground-roll wave. The result of frequency filtering is shown as follows: 
\begin{equation}
\begin{split}
&m = m_{low}+m_{high},\\
&m_{low} = g+r_{low},\\
&m_{high} = r_{high}
\end{split}
\end{equation}
where $r_{low}$ and $r_{high}$ represents the low- and high-frequency components of reflections. Since the $m_{high}$ contains no ground-roll contamination, this part stays unchanged during the whole process to protect the high frequency part of the reflections from being damaged. A multi-layer convolutional neural network (CNN) is applied afterwards to separate the low-frequency part of reflections and ground-roll wave apart. Considering that the high-frequency component of reflections $r_{high}$ tends to have similar patterns (such as angle, curvature) with low-frequency part $r_{low}$ in 2-D seismic image, so we assume that this kind of spatial similarity can be utilized as a guide and benefit the process for $r_{low}$ separation. As a result, we use the high-pass filtered result $m_{high}$ , which is the same as $r_{high}$ , together with $m_{low}$ as input of our convolutional neural network.

After the network is trained, we apply this processing to other mixture data of ground-roll and reflections, and the two outputs are the estimation of ground-roll and low-frequency components of reflections respectively.
By adding the high-frequency component of reflections which is not processed to the estimation of its corresponding low-frequency component, we then obtain the estimation of reflections. 

\subsection{CNN architecture and training approach}
\renewcommand{\figdir}{Fig} 


\begin{figure}
    \centering
    \includegraphics[width=0.9\linewidth]{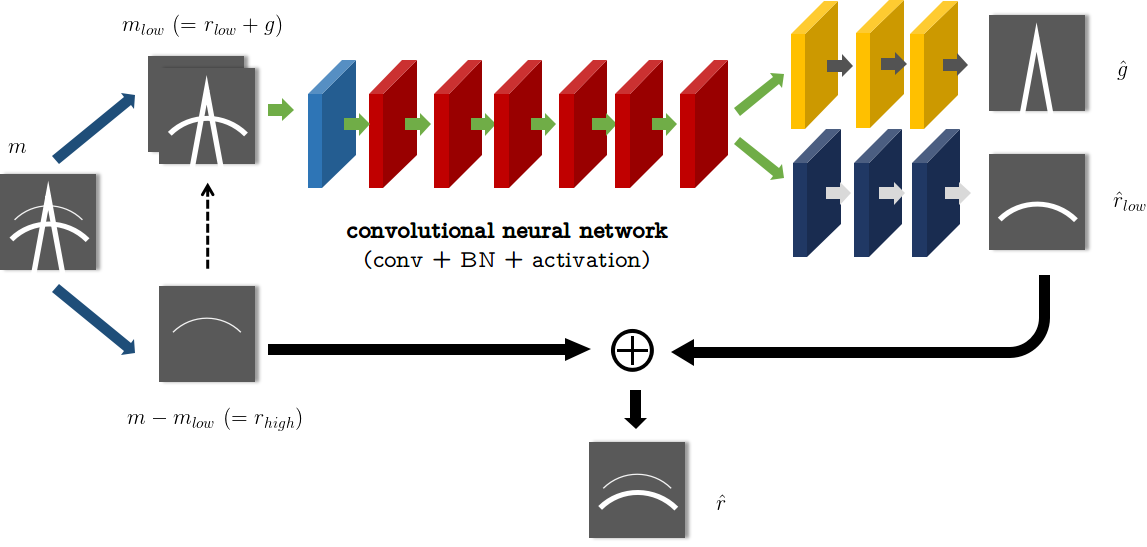}
    \caption{Flowchart of deep CNN based ground-roll separation method}
    \label{fig:fig1}
\end{figure}

In our method, the architecture of CNN model used to separate the low-frequency of mixture data is shown in Figure \ref{fig:fig1}.

The main components of this network are convolution operation (conv), batch normalization (BN) operation, and an activation function (LeakyReLU is used here). Convolution operation is to extract the spatial features of the input data. Batch normalization is commonly used in deep networks to accelerate the training process and prevent training from gradient vanishing or exploding by reducing the internal covariate shift in the hidden layers \cite{Ioffe2015Batch}. LeakyReLU is a modified version of the widely used ReLU activation to prevent the neurons from being dead in the negative axis. Apart from the first layer and the last two layers which give the outputs, each layer is consist of conv + BN + LeakyReLU. The first layer has no BN operation since our input signal is already normalized. The last two layers use the linear activation instead of LeakyReLU to generate outputs. As the figure shows, the first 7 layers of the network is the common path for both ground-roll output path and reflection output path. This common path is expected to learn the features from low-to high-level of mixture seismic data, and the following are 3 layers for each path to reconstruct the ground-roll and low-frequency part of reflections respectively. In addition, dropout layers are applied in the two output paths to prevent the model from overfitting. Each of the layers has the same number of 64 feature maps in our experiments, and convolution kernels are all of 3 $\times$ 3. The gradient descent training strategy used here is RMSprop, and the learning rate is set to 1e-4.

To train this CNN model using supervised learning method, we need to acquire some trace gathers containing ground-roll noise as our train data, and its denoised result with ground-roll noise delicately removed as well as the ground-roll itself as our reference labels. Then we use these train data and labels to form our train dataset. High-pass filter is applied to split the train data and corresponding label into high- and low-frequency components. The low-frequency components are utilized to train the model. The loss function for model training is as follows:

\vspace{-10px}

\begin{equation}
loss=||\hat{r_{low}}-r_{low}||_{2}^{2}+||\hat{g}-g||_{2}^{2}+||\hat{g}+\hat{r_{low}}-m_{low}||_{2}^{2}-\lambda||\hat{r_{low}}-\hat{g}||_{1}
\end{equation}

This loss function is composed of two parts, the first 3 terms in the loss function is similarity objective part, which is used to constraint the closeness of output and their corresponding reference label. The last term is discriminative objective part, this part is in order to make the two output components more distinguishable with each other. This kind of loss with discriminative part has been used in speech separation task successfully (\cite{huang_deep_2014}), the main target of it is to improve the similarity of the output to its reference label, while degrades the similarity between the output and the reference label of the other output. The parameter $\lambda$ is a tunable parameter which represents a trade-off of the importance between the similarity and discrimination.

\section{Experiments}

In order to illustrate the validity and effectiveness of proposed method, we conduct experiments on both synthetic and real seismic data. All experiments are conducted with Python 2.7 on a desk computer with Intel Core i7-7700K 4.20GHz CPU and Nvidia GeForce GTX 1080GPU with 8GB GPU memory. The CNN network in the method is implemented using the open framework PyTorch (version 0.4.0). 

\subsection{Synthetic data example}

In this section we evaluate the effectiveness of proposed method by experiments on synthetic seismic data. Firstly, we simulate both the ground-roll and the reflections. Synthesis of ground-roll wave is based on its properties of low frequency, dispersion, and low velocity. We use sweep signal to simulate the ground-roll wave, the sweep signal used here has the mathematical form as below:

\begin{equation}
g(t,x)=A(t,x)d(x)\sin(2\pi f(t)t)
\end{equation}

where $g(t,x)$ means the ground-roll value at time-space position $(t,x)$, and $A(t,x)$ represents the amplitude of the ground-roll signal train. In each trace $x_{i}$, amplitude $A(t,x_{i})$ has the form of a time window to determine the duration of the ground-roll wave. $d(x)$ is the degradation coefficient, and $f(t)$ is the sweep frequency to simulate the frequency dispersion effect of ground-roll wave.

The mathematical expression of $A(t,x)$ is as follows:

\vspace{-10px}

$$A(t,x)=
\begin{cases}
\mathrm{Tukeywin} (t-(x-x_c)/v)& \text{$(x-x_c)/v \leq t \leq (x-x_c)/v + T(x)$}\\
0 & \text{otherwise}
\end{cases}$$
where $\mathrm{Tukeywin}$ represents the Tukey window, which is a cosine-tapered time window of length $T(x)$. $x_c$ is the trace in the center, and $v$ is the apparent velocity of ground-roll, $T(x)$ is the duration of time window in trace number $x$. The degradation coefficient $d(x)$ has the form of 

\vspace{-10px}

$$d(x) = s^{x-x_c}$$
with $s$ a constant coefficient represents the scale or speed of degradation. In this experiment we use linear sweep signal which means $f(t)$ has the form :

\begin{equation}
f(t)=f_{b}+\frac{f_{e}-f_{b}}{2T(x)}t
\end{equation}
where $f_{b}$ and $f_{e}$ are the beginning and ending frequencies of sweep respectively, and $T(x)$ means the signal duration at trace number x \cite{Karsl2004Using}. 

In synthetic train dataset, we set $f_{b}=$5Hz and $f_{e}=$15Hz for the ground-roll wave, and apparent velocity of synthetic ground-roll varies from $\text{200m/s}$ to $\text{500m/s}$. The seismic wavelet used is Ricker wavelet with main frequency of 50 Hz, and linear event for direct wave and hyperbolic event with random amplitude and position for reflection wave are used to simulate the 2-D seismic structure of synthetic shot data. The synthetic data is shown in Fig. \ref{fig:fig2}.

\sideplot{fig2}{width=8cm}{Synthetic seismic gather containing ground-roll and reflections. (a) and (b) shows the $t-x$ and $f-k$ domain of the synthetic data. (c) shows one seismic trace where blue and red represents the synthetic seismic gather before and after ground-roll wave contamination.} 

Fig. \ref{fig:fig2} (a) is the $t-x$ domain and (b) the $f-k$ domain. Fig. \ref{fig:fig2} (c) shows one trace of synthesized reflections and mixture signal contaminated by ground-roll noise. Then we use the mixture of synthetic reflection data without ground-roll and synthetic ground-roll of different apparent velocities, time shift and amplitudes as raw train dataset of our method. Then we conduct a low-pass filtering with cut-off frequency of 25 Hz to split the low- and high-frequency of the mixture apart to train the CNN network. 

For the training stage, our training and validation dataset are generated using 10 synthetic seismic data which are different both in reflections and ground-rolls. Each of the synthetic data are of size $1000 \times 200$ (time samples $\times$ traces). Then all the synthetic data are cut into patches of size $64 \times 64$. Adjacent patches have a stride of 10. Through the cutting process, we generated our dataset used for training. The total training dataset used in the experiment has the size of  $13440\times 64\times 64\times 2$, 13400 is the number of patches, 64 is the patch size, and 2 for two channels with one low-frequency channel and one high-frequency channel. Validation set is split out from the whole dataset randomly with percentage of 0.1 to set the parameters and validate the method to make sure it works. We finally choose the parameter of $\lambda$ in the loss function as 0.001. After the parameters and network are set , we combine the training and validation set together for training process.  After the loss of network converges, we applied the method with trained CNN model to a synthetic mixture data where its reflections and ground-roll are not in the training set. The result and middle result are shown in Fig. \ref{fig:fig3}.
 
\sideplot{fig3}{width=8cm}{Synthetic test data results. (a) is the synthetic test data, (b) and (c) are the low-frequency component and high-frequency component split by the low-pass filter respectively. (d) and (e) are the outputs of the network, i.e. the separated ground-roll and low-frequency component of reflections respectively. (f) is the reconstructed reflections by adding (c) and (e) }

Fig. \ref{fig:fig3}(a) is the test mixture data. (b) and (c) are the low- and high-frequency part of (a), the high frequency part contains no ground-roll as illustrated before. (d) and (e) are the outputs of the CNN network, that is the low-frequency part of the reflections estimated and the ground-roll estimated. (e) shows the reconstructed reflection, that is the summation of (c) and (d). The above results illustrates the effectiveness of proposed method for separating ground-roll and reflections in the test synthetic seismic data. To show the performance of proposed method, we compare our result with that of $f-k$ dip filter method both in $t-x$ domain and $f-k$ domain as in Fig. \ref{fig:fig4}.

\sideplot{fig4}{width=7cm}{Comparison with $f-k$ dip filter result. (a) is the original synthetic data with ground-roll, (b) and (c) are the processed result of proposed method and $f-k$ dip filter. (d), (e), and (f) are the $f-k$ spectra corresponding to their left subfigures. The dip parameter has been tuned appropriately. }

 From the $f-k$ spectrum it is obvious that proposed method can effectively separate the reflections and ground-roll in low-frequency area, thus contribute to a better preservation and reconstruction of the reflection data. 

%
%
%
\subsection{Generalization test for trained model}

For the evaluation of deep learning models which use supervised training strategy with a certain train dataset, one important aspect is the generalization ability of the model, that is, the performance of the model when dealing with test data which is more different from the train dataset. Several experiments were conducted to test the generalization of our trained model. Firstly, considering that the synthetic ground-roll in our trainset has the same frequency dispersion (5Hz-15Hz), we test our trained model on the synthetic data with ground-roll has the frequency dispersion from 10Hz to 20Hz. Result is shown in Fig. \ref{fig:fig5_10-20Hz}.

\sideplot{fig5_10-20Hz}{width=7cm}{Generalization test on synthetic mixture data with ground-roll of different frequency dispersion (10Hz-20Hz). (a) and (b) are the original mixture data and reconstructed reflections, (c) and (d) are $f-k$ spectrum corresponding to their left $t-x$ domain seismic image}

From the $f-k$ spectrum it is shown that our model can still separate the ground-roll and reflections in the low-frequency part well. 

Another test is about the energy intensity of the ground-roll wave, i.e. the initial signal-to-noise ratio (SNR) if ground-roll is considered as noise. In the trainset we only used two different amplitude to synthesize the ground-roll, so in this experiment, a more strong ground-roll noise is applied in the mixture data for test (i.e. test data is of lower SNR). The result is shown in Fig. \ref{fig:gr_ratio3}. From Fig.\ref{fig:gr_ratio3} (c) it is clear that the ground-roll has more energy than before, (b) and (d) shows the effectiveness of our method on mixture data of different initial SNRs. 

Moreover, we vary the sampling rate of time and interval of traces (the time sampling rate and trace interval in trainset is 500Hz, 10m respectively). Fig. \ref{fig:t_down2} and \ref{fig:x_down2} shows the test results. Test data  has time sampling rate of 250Hz and trace interval 10m in Fig. \ref{fig:t_down2}, and 500Hz, 20m correspondingly in Fig. \ref{fig:x_down2}. From the results we can see that our method can be generalized to seismic data with sampling rate different from that of trainset. In $f-k$ spectrum it is obvious that the increase of trace interval  causes more serious aliasing effect, while proposed method still gives an acceptable result.

\sideplot{gr_ratio3}{width=7cm}{Generalization test on ground-roll with more energy.  (a) and (b) are the original mixture data and reconstructed reflections, (c) and (d) are $f-k$ spectrum corresponding to their left $t-x$ domain seismic image}

\sideplot{t_down2}{width=7cm}{Generalization test on data with time sampling rate decreased to 250Hz. (a) and (b) are the original mixture data and reconstructed reflections, (c) and (d) are $f-k$ spectrum corresponding to their left $t-x$ domain seismic image}

\sideplot{x_down2}{width=7cm}{Generalization test on data with trace interval increased to 20m. (a) and (b) are the original mixture data and reconstructed reflections, (c) and (d) are $f-k$ spectrum corresponding to their left $t-x$ domain seismic image}

Seismic data in the real world task is always with random noise. As a result, we tested our model on synthetic seismic data with additive Gaussian noise. The result is shown in Fig. \ref{fig:noise4}. From both $t-x$ domain and $f-k$ domain, we can see the ground-roll and reflections are well separated despite the low-quality of seismic image caused by additive random noise.

From all the generalization tests, we illustrate that after trained on a given dataset, our method can be adapted to other datasets even with different properties such as sampling rate, ground-roll frequency, and random noise level, while still achieving satisfactory results. A detailed quantitative analysis is shown in Table 1.

\sideplot{noise4}{width=7cm}{Generalization test on data with additive random noise, energy ratio of random noise before processing is 19.72\%, (a) and (b) are the original mixture data and reconstructed reflections, (c) and (d) are $f-k$ spectrum corresponding to their left $t-x$ domain seismic image}

Train data properties are shown in the first row, and experiment results on test data with different properties are shown below. The intensity of random noise is measured by the percentage of its energy in the energy of whole mixture signal. The SNRs before and after processing is shown in the table, with comparison to $f-k$ dip filter method. From the table we can see that results of our method have higher SNRs than that of $f-k$ dip filter. However the superiority decreases as the random noise becomes strong, because that in $f-k$ dip filter method, some of the random noise is also eliminated together with the ground-roll in the mask area. Moreover, as the $x$ sampling interval becomes 40m, the performance of proposed method tends to degrade, that means the generalization ability of the model still suffers some limitations.

\tabl{tblcomp}{Generalization test for test data of different properties, with comparison to $f-k$ dip filter result}
{
\scriptsize
\centering
\begin{tabular}{|c|c|c|c|c|c|c|c|}
\hline
& \makecell[c]{Ground-roll \\ dispersion} &\makecell[c]{Time sample \\ interval} & \makecell[c]{Trace interval} & \makecell[c]{random noise \\ (\%)} & \makecell[c]{Initial \\ (dB)} & \makecell[c]{$f-k$ dip \\ (dB)} & \makecell[c]{Proposed \\ (dB)} \\
\hline
Train & 5Hz-15Hz & 2ms & 10m & 0 & - & - & - \\
Test 0 & 5Hz-15Hz & 2ms & 10m & 0 & -11.07 & 10.77 & 15.15 \\
Test 1 & 10Hz-20Hz & 2ms & 10m & 0 & -11.13 & 11.13 & 16.32 \\
Test 2 & 5Hz-15Hz & 2ms & 10m & 0 & -20.61 & 1.65 & 14.83 \\
Test 3 & 5Hz-15Hz & 4ms & 10m & 0 & -11.07 & 10.82 & 16.59 \\
Test 4 & 5Hz-15Hz & 2ms & 20m & 0 & -11.07 & -0.09 & 16.78 \\
Test 5 & 5Hz-15Hz & 2ms & 30m & 0 & -10.95 & -4.31 & 17.19 \\
Test 6 & 5Hz-15Hz & 2ms & 40m & 0 & -11.09 & -7.73 & 14.33 \\
Test 7 & 5Hz-15Hz & 2ms & 10m & 3.53\% & -11.07 & 10.20 & 15.51 \\
Test 8 & 5Hz-15Hz & 2ms & 40m & 8.07\% & -11.09 & 8.88 & 11.21 \\
Test 9 & 5Hz-15Hz & 2ms & 40m & 13.22\% & -11.11 & 7.29 & 8.61 \\
Test 10 & 5Hz-15Hz & 2ms & 40m & 19.72\% & -11.14 & 5.74 & 6.54 \\
\hline
\end{tabular}
\label{comparetable}
}

\subsection{Land seismic record example}

Experiments have been also conducted on real seismic record. In the real land seismic experiments, our labels of separated reflections and ground-roll are obtained from the results processed manually using commercial software. The basic algorithm used in the software is “energy replacement method” ( \cite{Liang2017Energy}) The main idea of this method is based on the difference of ground-roll and effective signals. The ground-roll only occurs in the low frequency region, and has higher amplitude (more energy) than the reflections. The procedure of the algorithm is described as follows: 

1.Analysis of the spectrum of each single shot seismic data, and obtain the spectral band of ground-roll.
2.Split the seismic data into low and high spectral band using filtering method, then the high band contains only effective signals, while the low band contains effective signals and surface waves.
3.Conduct multi time-window rms (root mean square) amplitude gain analysis along t-axis on both low and high spectral band data, and calculate the gain factor in each time-window of each trace.
4.Compare the gain factor in low band and high band data, and use the larger gain factor of high band to substitute the smaller one correspondingly of the low band. Then use the processed gain factors to conduct an inverse gain calculation for the low-band data. Then the surface wave is suppressed.
5.Combine the low and high bands together to form the final result.

The method to generate the labels can satisfyingly suppress the ground-roll noise in the land seismic data, thus provide a proper label dataset for our model to learn from, but it need to design different time windows and analyze the the gain factors in each shot gather manually. The CNN based model is utilized to learn the features for the separation of reflections and ground-rolls from the manually processed labels, and produce an adaptive ground-roll separation result for each different shot gather. The training dataset has the size of $9088\times 64\times 64\times 2$, which means totally 9088 $64\times 64$ patches are used for training the network. The similarity loss and discriminant loss in the training stage is shown in Fig. \ref{fig:loss}. The training stage is stopped when both losses converge to a stable condition.

\sideplot{loss}{width=7cm}{Similarity loss and discriminant loss in the training stage for real land seismic data using proposed CNN based method}

After training finished, we then apply proposed method with trained network model to other seismic lines. Figures below shows our experiment result on real data. Fig. \ref{fig:zone1}, \ref{fig:zone2}, and \ref{fig:zone3} are 3 zones with far-, middle-, and near-offset to the source of surface wave, thus have different forms and shapes of their ground-roll noise. Fig. \ref{fig:zone1} (a) is the original data, (b) and (c) are the ground-roll noise and reflections separated by proposed method, for comparison, we also show the ground-roll separation results using $f-k$ dip filter in subfigure(d), and high-pass filter in (e), and the same with Fig. \ref{fig:zone2} and \ref{fig:zone3}. It can be seen that the ground-roll and reflections are well separated using proposed method, and the reflection events becomes more clear after separating ground-roll from it. The $f-k$ dip filter result contains strong ground-roll residue because the low frequency and low wave number part of ground-roll which is aliased with reflections cannot be removed using $f-k$ dip filter. The high-pass can remove the ground-roll well, yet it also loses the low-frequency information of reflections, which is not expected. 

Fig. \ref{fig:zone1fk}, \ref{fig:zone2fk}, and \ref{fig:zone3fk} shows the $f-k$ spectra corresponding to the results in Fig. \ref{fig:zone1}, \ref{fig:zone2}, and \ref{fig:zone3}. Fig. \ref{fig:zone1fk}(a), (b), and (c) are the spectra of original mixture data, separated ground-roll, and separated reflections respectively, (d) and (e) are the $f-k$ domain of $f-k$ dip filtered result and high-pass filtered result, and the same with Fig. \ref{fig:zone2fk} and \ref{fig:zone3fk}.  As shown in these figures, the proposed method can to some extent separate the overlapped frequency region of ground-roll and reflections, and thus preserves the reflections which are generally considered as the useful component. It is clear in $f-k$ domain that the high-pass filter removes low-frequency part of reflections when separating ground-roll, while the proposed method recovers the reflections in the overlapped region to some extent. Though $f-k$ dip filter can remove the low-frequency and high wavenumber (corresponding the low-velocity in $t-x$ domain) energy which is caused by ground-roll noise, it also shows incapability to deal with the overlapped region in $f-k$ domain of the two components. The above results illustrate that our proposed method is effective for ground-roll separation task in real seismic data.

\sideplot{zone1.jpg}{width=7cm}{Real shot data example in zone 1.  (a) is the original seismic data contaminated by ground-roll noise, (c) and (d) are the separated ground-roll and reflections respectively using proposed method. (e) and (f) are processed results using $f-k$ dip filter method and high-pass filter method respectively}

\sideplot{zone1fk}{width=7cm}{$f-k$ spectra of the real data results of zone 1. (a) $\sim$ (e) represents the $f-k$ spectra of the real mixture data, separated ground-roll, separated reflections, $f-k$ dip filter result, and high-pass filter result respectively. }

\sideplot{zone2.jpg}{width=7cm}{Real shot data example in zone 2.  (a) is the original seismic data contaminated by ground-roll noise, (c) and (d) are the separated ground-roll and reflections respectively using proposed method. (e) and (f) are processed results using $f-k$ dip filter method and high-pass filter method respectively}

\sideplot{zone2fk}{width=7cm}{$f-k$ spectra of the real data results of zone 2. (a) $\sim$ (e) represents the $f-k$ spectra of the real mixture data, separated ground-roll, separated reflections, $f-k$ dip filter result, and high-pass filter result respectively. }

\sideplot{zone3.jpg}{width=7cm}{Real shot data example in zone 3.  (a) is the original seismic data contaminated by ground-roll noise, (c) and (d) are the separated ground-roll and reflections respectively using proposed method. (e) and (f) are processed results using $f-k$ dip filter method and high-pass filter method respectively}

\sideplot{zone3fk}{width=7cm}{$f-k$ spectra of the real data results of zone 3.  (a) $\sim$ (e)  represents the $f-k$ spectra of the real mixture data, separated ground-roll, separated reflections, $f-k$ dip filter result, and high-pass filter result respectively. }

Moreover, we apply the trained CNN network to another land seismic data which is acquired in a different working area, so as to illustrate the generalization capability of the proposed model using real seismic data. In this experiment, the trained network is utilized directly to process the different data without any finetune.

Fig. \ref{fig:tianjia} shows the separation results. From Fig. \ref{fig:tianjia} it can be seen that the ground-roll in this working area has a relative weak energy, and differs from that of training data in morphological properties. However, the trained model still has the ability to separate the ground-roll from reflections to a certain extent, which indicates that our model has a degree of generalization ability in real land seismic data as well.

\section{Conclusion and future work}

In this paper, we proposed a novel approach to deal with the task of ground-roll wave attenuation in land seismic data via convolutional neural network. In this method we consider the task as a separation problem of two different kind of waves, and train an end-to-end CNN model with two outputs simultaneously to get the separated ground-roll and reflections. With the prior knowledge that ground-roll only contaminates a low and limited frequency band, our method only considers the separation of low-frequency part of seismic data. Clean high-frequency part is added directly to the estimated low-frequency part of reflections. Similarity objective of loss function controls the closeness between the output and reference label, while discriminative objective controls the difference between the separated outputs. Experiments on synthetic data and real land seismic data from oil company both have shown that the proposed method can separate the mixture data into ground-roll and reflections after training, and has generalization ability in a certain degree.

In our method, the training of CNN model requires data and corresponding label as our supervised train dataset, therefore the quality of the separation is related to the train dataset. Considering the independence of reflections to ground-roll, a more general objective to constraint the two outputs to be more independent or irrelevant can be applied to make our method less rely on the train dataset, which is what we will involve in in the future.

\sideplot{tianjia.jpg}{width=8cm}{ Ground-roll separation result of one selected seismic line in another working area directly using the trained model without finetune. The results show the generalization ability of the proposed CNN based model.}

\newpage
\bibliographystyle{seg} 
\bibliography{gr_sep_reference,example}
\end{document}